\documentclass[10pt]{article}
\usepackage{spconf, amsmath}
\usepackage{graphicx}
\usepackage{subfigure}
\usepackage{epstopdf}
\usepackage{amssymb}
\usepackage{bbm}
\usepackage{dsfont}
\usepackage{enumerate}
\usepackage[numbered]{algo}
\usepackage{url}

\usepackage[font=small,skip=0pt]{caption}
\def\Reals{{\mathbb R}}
\def\bQ{{\mathbf{Q}}}
\def\bq{{\mathbf{q}}}
\def\bmu{{\boldsymbol \mu}}
\def\bx{{\mathbf x}}

\def\bp{{\mathbf p}}
\def\balpha{{\boldsymbol \alpha}}
\def\br{{\mathbf r}}

\title{Coercive Region-level Registration for Multi-modal Images$^1$}
\name{Yu-Hui Chen$^{\star}$, Dennis Wei$^{\dagger}$, Gregory Newstadt$^{\ddagger}$, Jeffrey Simmons$^\P$ and Alfred Hero$^{\star}$}

\address{
  $^{\star}$ University of Michigan, Ann Arbor, MI USA \\
    $^{\dagger}$ IBM Watson Research Center, Yorktown Heights, NY USA \\
    $^{\ddagger}$ Google Inc., Pittsburgh, PA USA \\
    $^{\P}$ US Air Force Research Laboratory (AFRL), Dayton, OH USA}

\begin{document}
\ninept
\maketitle
\let\thefootnote\relax\footnotetext{\scriptsize $^1$~AOH would like to acknowledge financial support from USAF/AFMC grant FA8650-9-D-5037/04 and AFOSR grant FA9550-13-1-0043. MDG would like to acknowledge financial support from AFOSR MURI grant FA9550-12-1-0458.}

\begin{abstract}
We propose a coercive approach to simultaneously register and segment multi-modal images which share similar spatial structure. Registration is done at the region level to facilitate data fusion while avoiding the need for interpolation. The algorithm performs alternating minimization of an objective function informed by statistical models for pixel values in different modalities. Hypothesis tests are developed to determine whether to refine segmentations by splitting regions.  We demonstrate that our approach has significantly better performance than the state-of-the-art registration and segmentation methods on microscopy images.
\end{abstract}
\begin{keywords}
Image registration, Image segmentation, Multi-modality, Microscopy image, Hypothesis test
\end{keywords}
\section{Introduction}
\label{sec:intro}

This paper addresses the two problems of multi-modal image registration and image segmentation within a single framework. Multi-modal registration refers to registration of images acquired by different sensor/scanner types. It has been applied to many areas, e.g.~medical images, microscopy images, and radar images, to combine information from different modalities and provide more comprehensive understanding about the true object. Image segmentation, the partitioning of an image into meaningful regions, is an important step in image analysis and understanding. 

In this work, we focus on multi-modal registration and segmentation as applied to scanning electron microscope (SEM) images of materials; the methods to be discussed are equally applicable to other multi-modal images that share spatial structure. SEM techniques are widely used in materials science for material characterization, for example detection of defects that may cause fatigue when operating. Segmentation is of interest to map locations of \emph{grains}, uniform regions occupied by continuous crystal lattices, since grain structure is a principal factor in determining the properties of a polycrystalline material such metallic or ceramic materials~\cite{shah_effect_1988}. Multi-modal registration is desired because different scanning electron modalities carry complementary information \cite{wang_ebsd-afm_2005,nordmark_ebic_2008}. For example, Backscattered Electrons (BSE) provide information about topography and local fine-scale surface texture~\cite{goldstein_scanning_2003} while Electron Backscatter Diffraction (EBSD) measures crystal orientation which is useful in locating grains and grain boundaries~\cite{schwartz_electron_2009}. 

Multi-modal registration is made challenging by the fact that images from different modalities may have different resolutions, values that lie in different spaces (e.g.~scalars vs.~vectors), and different levels of distortion. In SEM for instance, these differences are due to different electron beam geometries, sensors, and recording electronics. Furthermore, there is often no complete forward model that jointly characterizes the multi-modal signals, nor a transformation model that adequately describes the distortion.  In these circumstances, pixel-level registration methods~\cite{maes_multimodality_1997,rueckert_nonrigid_1999,szeliski_spline-based_1997}, i.e., those that establish correspondences between pixels, usually resort to interpolation, a somewhat ad-hoc solution that may bias the resulting images toward excessive smoothness. On the other hand, segmentation of multi-modal images, if done independently for each modality using existing methods~\cite{chuang_automated_2008,nock_statistical_2004,schwartz_electron_2009}, may fail due to low contrast in some modalities and face difficulties in identifying correspondences between segmented regions from different modalities. 

In this work, we propose a coercive region-level approach to simultaneously register and segment images of different modalities that share similar spatial structure. The algorithm is initialized by segmenting one image by a standard method and coarsely mapping the result onto the other image. Then the two images are registered at the region level and further segmented through alternating minimization of a statistically-based objective function. There are several advantages of our approach. First, the region-level approach is free of pixel value interpolation and its inherent assumptions. Second, it takes advantage of modalities with better discriminative power, improving the overall segmentation result. The approach also preserves region correspondences to facilitate data fusion in~\cite{piella_general_2003,wan_segmentation-driven_2009}. Lastly, both registration and refinement of segmentation are driven by statistical models. In particular, we propose hypothesis tests to detect boundaries that are missed by the initial segmentation due to low contrast.

The paper is organized as follows. In Section~\ref{sec:system_framework}, we describe the objective function, statistical models for data from different modalities, and optimization methods for the region-level registration algorithm. In Section~\ref{sec:hypotest_missingedge}, we focus on hypothesis testing for detecting missing boundaries. Section~\ref{sec:experiment} shows experimental results for synthetic and real materials images and compares several different approaches. Section~\ref{sec:conclusion} concludes this paper.

\section{Algorithm Framework and Models}
\label{sec:system_framework}
\subsection{Objective Function}
We assume that there are two images from different modalities. A pixel location $(x,y)\in \mathcal{I}_i, i\in\{1,2\}$ is represented as vector $\bp$, where $\mathcal{I}_i$ is the spatial domain for the $i$-th modality. The pixel value at $\bp$ is given by a function $I_i(\bp)$. Note that the values $I_{1}(\bp_{1})$ and $I_{2}(\bp_{2})$ may lie in different spaces. The region-level registration problem is to find partitions of $\mathcal{I}_i$, $S_i=\{R_{i_j}\}_{j=1}^{N_i}$, where each segment $R_{i_j}$ is a collection of connected pixel locations and $N_i$ is the number of segments, to minimize the following objective function:
\begin{equation}
\label{eq:ObjFunction}
U(S_1,S_2)=J(S_1, I_1)+J(S_2,I_2)+\lambda D(S_1, S_2),
\end{equation}
where $J(S_i,I_i)$ is the intra-modal energy function that measures how well the segmentation fits the image data and $D(S_1, S_2)$ is the inter-modal energy function that coerces the segmentation results 
to be topologically similar, motivated by the fact that they share the same underlying physical structure. The parameter $\lambda$ controls the relative importance of the two terms. 

In this paper, we define the inter-modal energy $D(S_1,S_2)$ to be the number of segment boundaries that are present in one modality but not the other. This number is easily tracked because our algorithm maintains tight correspondences between segments in the two images. More generally, segment structure can be represented by a connected adjacency graph and the inter-modal energy can be any function which measures the topological distance between two graphs. The intra-modal energy $J(S_i,I_i)$ is defined by the statistical models described in the following subsection.

\subsection{Statistical Models for Pixel Values}
\label{subsec:statModel}
In the materials context, each segment $R_{ij}$ corresponds to a grain, a continuous crystal lattice. Motivated by this, we assume that the observed values within a segment are similar and can be modeled by i.i.d. random variables following a distribution with the same parameters. In the sequel, the image modality subscript $i$ is suppressed for simplicity. Let the probability density function (PDF) of the distribution for one modality be denoted by $f(I(\bp)|\balpha)$, where $\balpha$ represents the parameters specifying the model. The intra-modal energy function in (\ref{eq:ObjFunction}) given a set of segments $S=\{R_1, R_2,...,R_N\}$ is defined as: 
\begin{equation}
\label{eq:region_comp_obj}
J(S, I)=\sum_{j=1}^N \left[ -\sum_{\bp\in R_j}\log{f(I(\bp)|\hat{\balpha}_j)} +\epsilon\int_{\partial R_j}dl \right]
\end{equation}
where $\partial R_j$ is the boundary of region $R_j$ with counter-clockwise definition and $\hat{\balpha}_j$ is the maximum-likelihood (ML) estimate for the parameters of region $R_j$. The first term is the negative log-likelihood of observations which penalizes grain inhomogeneity and the second term penalizes the boundary length, where $\epsilon$ controls the level of smoothness. 

In this paper, we consider the EBSD and BSE images of one section of a material as our input. Note that other image types can be used directly given properly defined statistical models. For BSE images, since the pixel values are scalars, the intensities in the same grain region are modeled by a univariate Gaussian $\mathcal{N}(\mu_j, \sigma_j^2)$, where $\mu_j, \sigma_j^2$ are the mean and variance of $R_j$. Notice that $\mu_j$ and $\sigma_j^2$ are unknown parameters to be estimated from image data. 

For EBSD images, the pixel values characterize the local crystal orientation, which can be represented by Euler angles~\cite{eberly_euler_2008}, Rodrigues vectors~\cite{morawiec_rodrigues_1996} or quaternions~\cite{altmann_rotations_2005}. We choose the unit-quaternion representation, i.e.~a $\bq\in S^3$, the $3$-dimensional unit sphere in $\Reals^4$. This allows use of the von Mises-Fisher (VMF) distribution in directional statistics~\cite{mardia_directional_1999}, a natural generalization of the multivariate Gaussian distribution to the sphere $S^{(p-1)}\subset \Reals^p$ (here $p=4$). However, the VMF distribution cannot be used directly due to crystal symmetry, which causes there to be more than one quaternion representation corresponding to a single crystal orientation. The ambiguity in representation may lead to a large artificial diversity of pixels within the same grain, resulting in an over-segmented result. To cope with this problem, we have proposed a VMF mixture distribution model which accounts for symmetry in our previous work~\cite{chen_parameter_2015}. To briefly describe the model, let $\mathcal G$ be a group of quaternion matrices $\{\bQ_1, \ldots, \bQ_M\}$ which define the symmetry actions that map a quaternion $\bq$ to its symmetric equivalents. The PDF of the pure VMF distribution is $\phi(\bx;\bmu,\kappa)=c_p(\kappa)\exp(\kappa\bmu^T\bx)$, where $\bx,\bmu\in S^{(p-1)}$, $\bmu$ is the mean direction, $\kappa$ is the concentration parameter, $c_p(\kappa) = \frac{\kappa^{p/2-1}}{(2\pi)^{p/2}I_{p/2-1}(\kappa)}$ and $I_p(.)$ is the modified Bessel function of the first kind with order $p$. The density function of the VMF mixture distribution is then given by
\begin{equation}
\label{eq:mixture_density}
f(\bx;\bmu,\kappa)=\sum_{m=1}^M\frac{1}{M}\phi(\bx;\bQ_m\bmu, \kappa)
\end{equation}
The parameters $\bmu$ and $\kappa$ can be estimated from image data through the Expectation-Maximization algorithm derived in~\cite{chen_parameter_2015}.

\subsection{Optimization}
We minimize the objective function (\ref{eq:ObjFunction}) by alternately fixing $S_1^{(k)},S_2^{(k)}$ and solving (\ref{eq:ObjFunction_part1}) and (\ref{eq:ObjFunction_part2}),

\begin{align}
\label{eq:ObjFunction_part1}
S_2^{(k+1)} &= \arg\min_{S}J(S,I_2)+\lambda D(S_1^{(k)},S)\\
\label{eq:ObjFunction_part2}
S_1^{(k+1)}&=\arg\min_{S} J(S,I_1)+\lambda D(S,S_2^{(k+1)})
\end{align}
where $k$ is the iteration index. Typically 2--3 iterations suffice. 

To initialize the algorithm, the initial segmentation of the first modality, $S_1^{(0)}$, is obtained by using a suitable image segmentation method. For example, the Voronoi-based method in~\cite{bachmann_grain_2011} can be applied to EBSD images and the Stabilized Inverse Diffusion Equation method to BSE images~\cite{chuang_automated_2008}. Since EBSD data provides crystal orientation which defines grain regions more accurately, we choose to start with EBSD segmentation in this paper. Next, to account for global misalignment and any resolution difference between the modalities, we determine an affine transformation by treating the material sample as a binary image and registering its outer boundary from one modality to the other using the Elastix toolbox~\cite{klein_elastix:_2010}. The transformation is then used to map $S_1^{(0)}$ onto the other modality, yielding the initial segmentation $S_2^{(0)}$. Note that due to localized distortions between the modalities, the initial segmentation $S_2^{(0)}$ may be misaligned with the image values as shown in Fig.\ref{fig:misalignment:a} and therefore needs to be registered. 

Optimizing (\ref{eq:ObjFunction_part1}) and (\ref{eq:ObjFunction_part2}) is done in two steps. The first step is to consider splitting regions in the current segmentation by adding boundaries. In Section~\ref{sec:hypotest_missingedge}, we propose a hypothesis testing approach for this purpose based on the statistical model \eqref{eq:region_comp_obj}. The second step is to register the misaligned boundaries. Due to the fact that adjusting boundary positions does not change the topology of the segment structure, the inter-modal energy function $D(S_1,S_2)$ is not changed in this step, reducing (\ref{eq:ObjFunction_part1}) and (\ref{eq:ObjFunction_part2}) to the intra-modal energy function $J(S,I)$ alone, which is given by our statistical model. We use the Region Competition algorithm~\cite{zhu_region_1995} to minimize $J(S,I)$. This algorithm applies gradient descent to move pixels comprising the boundaries $\partial R_j$ along their respective normal directions. There are two forces driving the movement corresponding to the two terms in \eqref{eq:region_comp_obj}: the statistics force which comes from the distribution model for the pixel values, and the smoothing force which drives the boundary to have smaller curvature. More details are given in \cite{zhu_region_1995}.

\section{Hypothesis Tests for Missing Boundaries}
\label{sec:hypotest_missingedge}
\begin{figure}
  \centering
  \subfigure[Misalignment]{
    \label{fig:misalignment:a} 
    \includegraphics[width=0.95in]{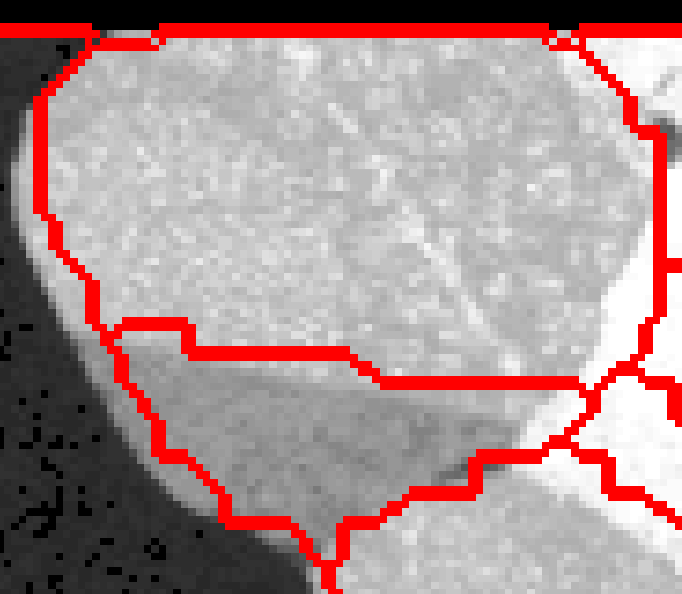}}
    \hspace{7mm}
  \subfigure[Missing Edge]{
    \label{fig:misalignment:b} 
    \includegraphics[width=0.95in]{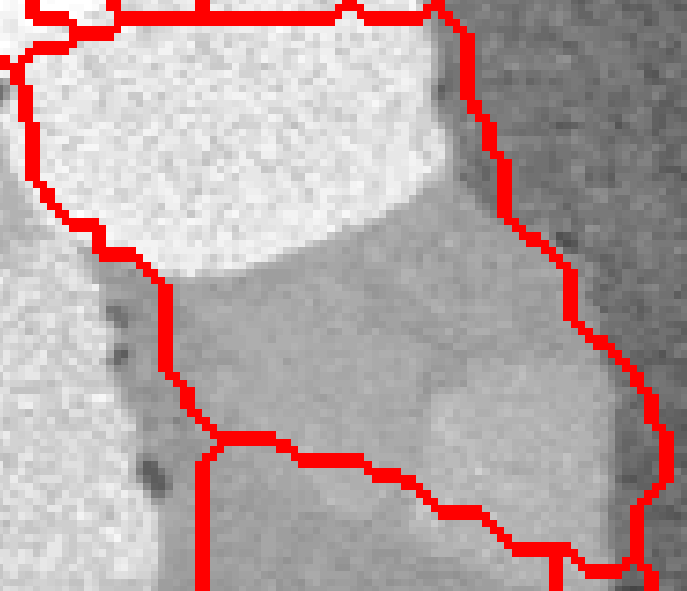}}
  \caption{
  (a) shows a misaligned 
  boundary while 
  (b) indicates a missing boundary. 
  Notice that in both 
  situations, pixel values within the regions outlined in red are multi-modally distributed.}
  \label{fig:MisalignmentAndMissingEdge} 
\end{figure}

This section elaborates upon the first step in solving \eqref{eq:ObjFunction_part1} and \eqref{eq:ObjFunction_part2}, namely hypothesis testing to determine whether a region $R\in S$ should be split into two based on the observed image values. We refer to this as the missing boundary problem. Recall that $R$ may come from the initial segmentation result from another modality and may not fit the current image data. Figure \ref{fig:MisalignmentAndMissingEdge} shows examples of misalignment and a missing boundary. One can see that both of the situations have multi-modal distributions of pixel values within the initially defined regions but only Fig.\ref{fig:misalignment:b} shows a missing edge that should be identified. Therefore, a region $R$ is declared as having a missing boundary if and only if it satisfies the following two conditions: (1) The pixel values are multi-modally distributed. (2) The multi-modal distribution is unlikely to be caused by misalignment. We develop two hypothesis tests for the two criteria. The first hypothesis test uses the Generalized Likelihood Ratio Test (GLRT)~\cite{neyman_problem_1992} to test whether the pixel values are multi-modally distributed. The second hypothesis test differentiates misalignment from a missing boundary. 

\setlength{\dbltextfloatsep}{1pt}

\subsection{Hypothesis Test for Multi-Modality}
Recall from Section~\ref{subsec:statModel} that 
the set of pixel values within a region $R$ are modeled by a distribution $f(I(R); \balpha)$ with unknown parameters $\balpha$, where $I(R)=\{I(\bp)\}_{\bp\in R}$ are the observed pixel values in $R$. Assume there exists a boundary $\psi$ which partitions $R$ into two sub-regions $R_+, R_-$ with parameters $\balpha_{+}, \balpha_{-}$. The two hypotheses are $H_0$: region $R$ is indeed a single region, i.e. $\balpha_+ = \balpha_-$; and $H_1$: $R$ consists of two regions. The GLRT has the following form:

\begin{eqnarray}
\label{eq:GLRT_formula_ourcase}
\log{\Lambda_{GLR}} &=& \log{\max_\psi \frac{\max_{\{\balpha_+, \balpha_-\}}f(I(R); \balpha_+, \balpha_-, \psi)}{\max_{\{\balpha_+=\balpha_-\}}f(I(R); \balpha_+, \balpha_-, \psi)}} \nonumber \\
&=& \max_\psi \sum_{p\in R_+}\log{f(I(p)|\hat{\balpha}_+)}+\sum_{p\in R_-}\log{f(I(p)|\hat{\balpha}_-)} \nonumber \\
&&-\sum_{p\in R}\log{f(I(p)|\hat{\balpha})} \gtrless_{H_0}^{H_1} \lambda
\end{eqnarray}
where $\hat{\balpha}, \hat{\balpha}_+, \hat{\balpha}_-$ are the ML estimates of the parameters under the null and alternative hypotheses and $\lambda$ is the coefficient in (\ref{eq:ObjFunction}). The GLRT can be viewed as a trade-off between the improvement in the intra-modal energy and the penalty of $\lambda$ paid in the inter-modal energy when inserting a boundary. The boundary length penalty is neglected here for simplicity but can be included easily. 

In the following subsections, we derive the GLRT for univariate Gaussian and VMF distributions given the boundary $\psi$. We only discuss the equal variance (concentration parameter) case due to the paper length constraint. These expressions supply the objective function, denoted as $\Lambda_{GLR|\psi}$, to be maximized with respect to $\psi$ in \eqref{eq:GLRT_formula_ourcase}. We use the Region Growing algorithm~\cite{zhu_region_1995} to locate the optimal boundary $\psi$. The algorithm partitions a region starting 
from two seed pixels and greedily adds neighboring pixels until all pixels in the region are chosen.

\subsubsection{Multi-modality Test for Univariate Gaussian Distribution}
\label{sec:unitest_Gau}
The GLR $\Lambda_{GLR|\psi}$ given boundary $\psi$ for testing mean equality between two Gaussian distributions has the following form~\cite{seize_students_1977}:

\begin{equation}
\label{eq:GLRT_uniGau_caseI_formula}
\Lambda_{GLR|\psi} = \left(\frac{\hat{\sigma}_0^2}{\frac{n_+}{n}\hat{\sigma}_+^2+\frac{n_+}{n}\hat{\sigma}_-^2}\right)^{n/2},
\end{equation}
where $\hat{\sigma}_0^{2}, \hat{\sigma}_+^{2}, \hat{\sigma}_-^{2}$ are the ML estimators of the variances under the null and alternative hypothesis and $n,n_+,n_-$ are the numbers of pixels in regions $R,R_+,R_-$. The optimization of the boundary $\psi$ then takes the form 

\begin{equation}
\label{Objective_uniGau_caseI_formula}
\begin{split}
\hat{\psi}&= \arg\min_{\psi} \sum_{\bp\in R_+} (I(\bp)-\hat{\mu}_+)^2 + \sum_{\bp\in R_-}(I(\bp)-\hat{\mu}_-)^2
\end{split}
\end{equation}
where $\hat{\mu}_{\pm}$ is the ML estimate of the mean in $R_{\pm}$.

\subsubsection{Multi-modality Test for von Mises-Fisher Distribution}
\label{sec:unitest_VMF}
The VMF mixture distribution is reduced to single VMF through transoforming the samples by the symmetry operator towards the mean direction estimated by the EM algorithm.
According to the derivation of the ML estimators in~\cite{dhillon_modeling_2003}, $\Lambda_{GLR|\psi}$ has the following form:

\begin{equation}
\label{eq:GLRT_VMF_caseI_formula}
\Lambda_{GLR|\psi}=\frac{c_p(\hat{\kappa}_1)^{n}}{c_p(\hat{\kappa}_0)^n}\exp{(\hat{\kappa}_1(\|\br_+\|+\|\br_-\|)-\hat{\kappa}_0\|\br_0\|)}
\end{equation}
where $\br_+ = \sum_{\bp\in R_+} I(\bp), \br_- = \sum_{\bp\in R_-} I(\bp), \br_0=\br_++\br_-$ and $\hat{\kappa}_1 = A_p^{-1}((\|\br_+\|+\|\br_-\|)/n), \hat{\kappa}_0=A_p^{-1}(\|\br_0\|/n)$, $A_p(x)=I_{p/2}(x)/I_{p/2-1}(x)$. The optimization over $\psi$ is 
\begin{equation}
\label{eq:Objective_VMF_caseI_formula}
\begin{split}
\hat{\psi} &= \arg\max_{\psi} n(\log{c_p(\hat{\kappa}_1)}+\hat{\kappa}_1A_p(\hat{\kappa}_1)) \\
&= \arg\max_{\psi} \|\sum_{\bp\in R_+}I(\bp)\|+\|\sum_{\bp\in R_-}I(\bp)\|
\end{split}
\end{equation}
The last equality comes from the fact that $n(\log{c_p(x)}+xA_p(x))$ and $A_{p}(x)$ are monotonically increasing functions of $x$.

\subsection{Hypothesis Test for Misalignment}
For regions that pass the previous multi-modality test ($H_{1}$ declared in \eqref{eq:GLRT_formula_ourcase}), we perform a second hypothesis test to determine whether the multi-modal distribution is due to $H_0$: boundary misalignment, or $H_1$: a missing boundary. Since in most cases, misalignment causes only a small portion of pixels to differ from the majority, one naive test is to set a threshold on the ratio of the size of the smaller region to the whole region:

\begin{equation}
\label{missingEdge_test_formula}
T=\frac{\min\{|R_+|, |R_-|\}}{|R|}\gtrless_{H_0}^{H_1}\eta
\end{equation}

\setlength{\textfloatsep}{1pt}
\begin{figure}
  \centering
  \subfigure{
    \label{fig:grain_model:a} 
    \includegraphics[width=1.00in]{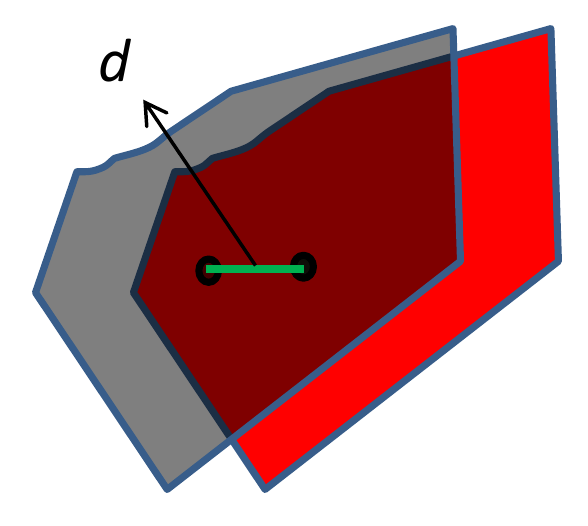}}
    \hspace{10mm}
  \subfigure{
    \label{fig:grain_model:b} 
    \includegraphics[width=1.00in]{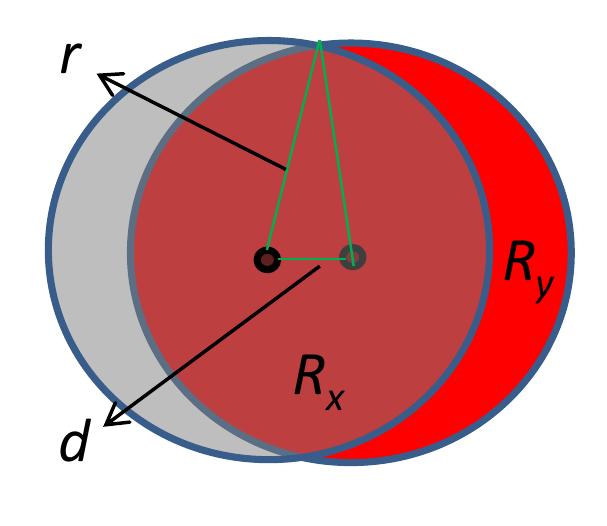}}
  \caption{
  Misalignment caused by displacement for (a) a realistic region shape; (b) a 
  simplified circular 
  model.}
  \label{fig:grain_model} 
\end{figure}
However, since region size can vary over several orders of magnitude, the same absolute amount of misalignment (in pixels) can result in very different size ratios, making it hard to set a universal threshold. Therefore, we propose an adaptive threshold which incorporates region size. Boundary misalignment is modeled by a displacement in position (see Fig.\ref{fig:grain_model:a}), where the displacements $(d_x,d_y)$ are bivariate Gaussian with zero mean and covariance $\mathbf{\Sigma}_d=\sigma_d^{2} \mathbf{I}_2$ and $\mathbf{I}_2$ is the $2\times2$ identity matrix. For simplicity, the region is modeled as a circle with radius $r$ (Fig.~\ref{fig:grain_model:b}), where $r=\sqrt{|R|/{\pi}}$ is the equivalent radius of region $R$. Based on these assumptions, the test statistic in (\ref{missingEdge_test_formula}) can be formulated as the following function of $d=\sqrt{x^2+y^2}$ given $r$:
\begin{equation}
\begin{split}
&T=f_r(d) = 1-\frac{2}{\pi}\arccos(\frac{d}{2r})+\frac{d}{\pi r^2}\sqrt{r^2-d^2/4} \nonumber \\
&\Rightarrow d=f_r^{-1}(T)\gtrless_{H_0}^{H_1} f_r^{-1}(\eta)=\eta'
\end{split}
\end{equation}
The second line follows because $f_{r}$ is an increasing function. Since the displacement $d$ follows a Rayleigh$(\sigma_{d})$ distribution, given the user specified false positive rate $\alpha$, we set 
\begin{equation}
\begin{split}
&\alpha = P(d>\eta' | H_0)
= Q(\eta') = P(T > f_r(\eta')|H_0) \nonumber \\
&\Rightarrow \eta = f_r(Q^{-1}(\alpha))
\end{split}
\end{equation}
where $Q(.)$ is the Rayleigh tail distribution. As a result, the threshold is adaptively determined by $\alpha$ and the equivalent radius $r$.

\vspace{-8pt}
\section{Experiments}
\label{sec:experiment}
\subsection{Boundary Detection Accuracy on 
Simulated Data}
In this section, we compare grain boundary detection performance on simulated EBSD and BSE images using three different approaches: A. Segment the BSE and EBSD images separately by suitable existing segmentation algorithms \cite{nock_statistical_2004,bachmann_grain_2011}; B. Segment EBSD and register the boundaries onto BSE using a B-Spline deformation model and the mutual information criterion~\cite{maes_multimodality_1997,rueckert_nonrigid_1999}; C. The proposed coercive registration/segmentation algorithm with $\lambda=0.15, \epsilon=25, \alpha=0.05$.

The grain shapes in the testing data are taken from real microscopy images downloaded from BlueQuartz~\cite{_bluequartz_2003} and segmented by their Dream3D toolbox. For each slice, some of the grains are randomly selected and displaced to produce boundary misalignment according to the Gaussian displacement model with $\sigma_d=3$ (pixels). This creates the ground truth boundaries for evaluation. The pixel values for BSE and EBSD are generated from Gaussian and VMF distributions with random mean and variance/concentration for each grain region.

To evaluate the boundary detection accuracy, we use the ``overlapping rate". Let $B(w)$ be the set of boundary pixel locations with boundary width $w$, which is obtained by image dilation with filter disk radius $w/2$. The overlapping rate is defined as $O(w)=|B_T(w)\cap \hat{B}(w)|/|B_T(w)\cup \hat{B}(w)|$, where $B_T(w),\hat{B}(w)$ are the ground truth and estimated boundary.

\begin{figure}
\begin{center}
\includegraphics[width=5.6cm]{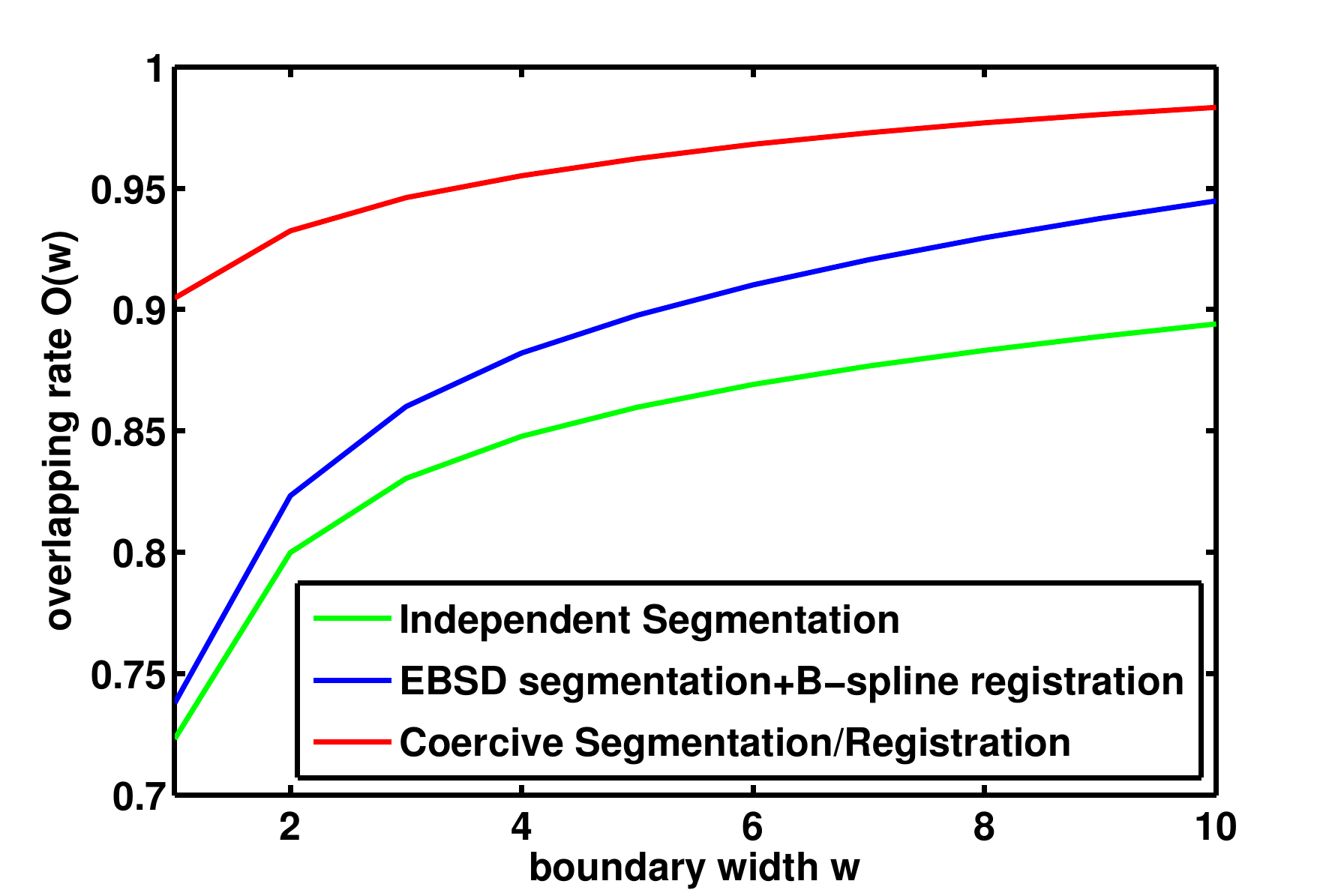}
\end{center}
\caption[]{The proposed coercive approach (red line) has much higher boundary overlapping rate than other approaches since it is able to detect missing boundaries and register misaligned boundaries.}
\label{fig:CompareRateAll}
\end{figure}

Figure \ref{fig:CompareRateAll} shows the overlapping rate of the three approaches for different boundary widths. Independent segmentation has the worst performance since it does not make use of shared sub-structure between modalities. With B-spline registration, there is some improvement but it is still not satisfactory, especially for small $w$. The proposed coercive registration approach with hypothesis testing is able to accurately register misaligned boundaries and detect missing edges. Therefore, it has much better boundary detection performance.

\setlength{\textfloatsep}{1pt}
\subsection{Results on Real Microscopy Data}
\begin{figure}
\begin{center}
\includegraphics[width=4.9cm]{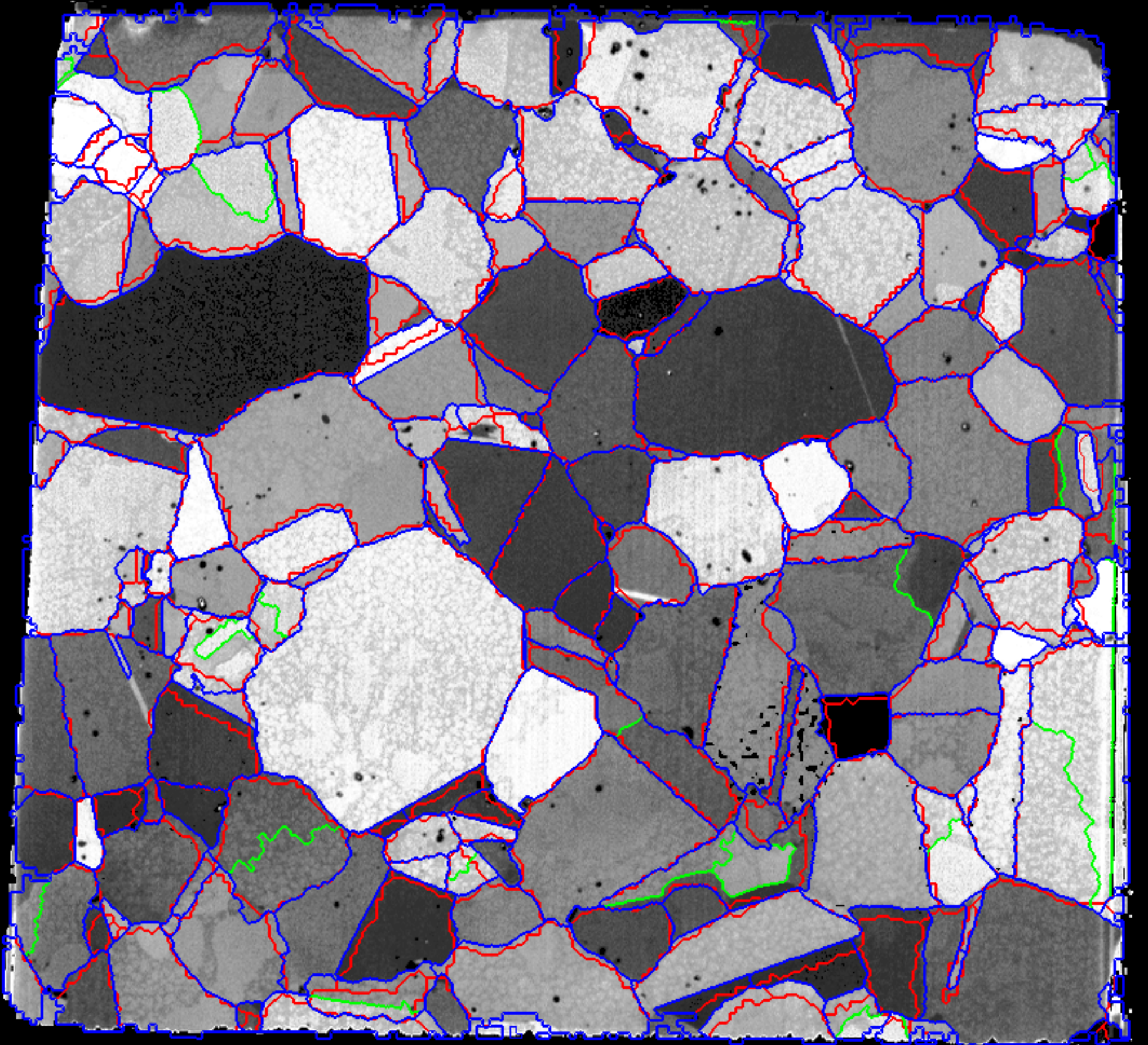}
\end{center}
\caption[]{The registered boundaries (blue lines) fit the BSE image values much better than the initial boundaries from EBSD (red lines). The proposed approach is also able to detect and locate missing boundaries within grain regions (green lines).}
\label{fig:IN100_result}
\end{figure}

We apply the proposed method to the IN100 data set which contains 170 slices of EBSD and BSE images of a Ni-base alloy. Figure \ref{fig:IN100_result} shows one registration/segmentation result overlaid on the BSE image. The red lines are the initial boundaries obtained by the EBSD segmentation and affine-transformed to match BSE. The blue lines are the realigned boundaries and the green lines are the missing boundaries detected by the hypothesis tests. The initial red lines are misaligned with the BSE image values but are corrected by our registration algorithm. Using statistical hypothesis tests, we are also able to detect and locate missing boundaries in some grain regions. These results in real data demonstrate that the proposed approach can accurately register boundaries and segment grain regions at the same time.

\setlength{\textfloatsep}{1pt}

\vspace{-8pt}
\section{Conclusion}
\vspace{-3pt}
\label{sec:conclusion}
In this work, we proposed a coercive registration/segmentation algorithm for multi-modal images. The algorithm alternately utilizes information from one modality to help segment the image in the other modality, resulting in significant performance improvement in both modalities. The proposed hypothesis test based on statistical models of pixel values can accurately detect and locate missing boundaries between regions. Furthermore, our approach identifies and preserves all of the correspondences between regions in different modalities, which is important for fusing information after registration. The experiment results on simulated and real microscopy images show that our approach is able to effectively correct misaligned grain boundaries and detect missing boundaries within grain regions.

\bibliographystyle{IEEEbib}
\bibliography{ICIP2015}

\end{document}